\documentclass{article}

\usepackage{url}
\usepackage{authblk}
\usepackage{tikz}
\usepackage{dsfont}
\usepackage{amsmath}
\usepackage[acronym, nonumberlist]{glossaries}
\usepackage{microtype}
\usepackage{graphicx}
\usepackage{subfigure}
\usepackage{booktabs} 
\usepackage{hyperref}

\title{Model Agnostic Combination for Ensemble Learning}
\author{Ohad Silbert}
\author{Yitzhak Peleg}
\author{Evi Kopelowitz}
\affil{Philips Algotec}

\newcommand{\xgboost}{XGBoost }

\newacronym{DNN}{DNN}{Deep Neural Network}
\newacronym{RELU}{ReLU}{Rectified Linear Unit}
\newacronym{MAC}{MAC}{Model Agnotsic Combination}
\newacronym{SGD}{SGD}{Stochastic gradient descent}
\newacronym{CT}{CT}{Computed Tomography}

\begin{document}
\maketitle

\begin{abstract}
Ensemble of models is well known to improve single model performance. 
We present a novel ensembling technique coined \gls*{MAC} that is designed to find the optimal function for combining models while remaining invariant to the number of sub-models involved in the combination. 
Being agnostic to the number of sub-models enables addition and replacement of sub-models to the combination even after deployment, unlike many of the current methods for ensembling such as stacking, boosting, mixture of experts and super learners that lock the models used for combination during training and therefore need retraining whenever a new model is introduced into the ensemble. 
We show that on the 
\href{https://www.kaggle.com/c/rsna-intracranial-hemorrhage-detection}{Kaggle RSNA Intracranial Hemorrhage Detection challenge}, \gls*{MAC} outperforms classical average methods, demonstrates competitive results to boosting via \xgboost for a fixed number of sub-models, and outperforms it when adding sub-models to the combination without retraining.
\end{abstract}

\section{Introduction}
\label{sec:introduction}
In machine learning, an ensemble is a collection of sub-models that are combined together to form a single predictive model \cite{when_networks_disagree, ensemble_methods_review_book}. 
An ensemble performs better than a single model \cite{when_networks_disagree, ensemble_methods_review_book, ju2017relative_superlearner} and has become a common practice in most machine-learning competitions \cite{kagglewriteups, xgboost, he2015deep_resnet, simonyan2014deepilsvrc2014, NIPS2012_imegenet}. 
Many ensembling methods exist in the literature, such as boosting \cite{boosting}, bagging \cite{bagging}, stacking \cite{stacking}, mixture of experts \cite{mixer_of_experts}, super learners \cite{ju2017relative_superlearner} and many more. 
Some of the methods focus on maximizing the sub-model diversity in the ensemble, usually by sampling the training set \cite{wen2020batchensemble, ensemble_methods_review_book, when_networks_disagree}. 
The combination of the sub-models is then performed by a simple majority voting or averaging which stabilize the final prediction with respect to the sub-models predictions variance. 
In some cases, like in boosting \cite{boosting} or mixture of experts \cite{mixer_of_experts}, a more complex combination exploits the performance differences between the sub-models in order to enhance the strength of each sub-model. 

Although sophisticated ensembling methods, such as boosting or mixture of experts, outperform simple and weighted averages, their primary disadvantage is their tight relation to the specific sub-models they were trained on. 
Adding or replacing even a single sub-model in the ensemble will require retraining which is often computationally expensive. 
On the other hand, combination methods that are based on averaging are inherently invariant to the specific sub-models in the combination and are therefore flexible to adding, removing or replacing sub-models without changing the combination scheme itself.

Multiple average based combination functions exist, for example: the harmonic mean is more suitable when averaging rates and the geometrical mean is often used with exponential variables like population growth. 
One way to improve an average based combination is to find the optimal combination function for a given task. 
In this paper we aim to maintain the property of being agnostic to the number of sub-models and their detailed differences and generalize the average based combination by building a framework that searches for the optimal combination function in a supervised manner. 
Unlike conventual ensembling techniques, like boosting and stacking, in this framework, coined \gls{MAC} (Model Agnostic Combination), the sub-model are not weighed in the combination with respect to the overall sub-model performance. 
Instead, each prediction, coming from any sub-model, is transformed to a learned latent space for which the combination function is optimized. 
The unique structure of \gls{MAC} is therefore agnostic to specific sub-models used for optimization, and can combine predictions with a different number of sub-models without the need to re-optimize the combination function. 

To test this method, we evaluate the performance of simple average, \xgboost \cite{xgboost} and \gls{MAC} on the \href{https://www.kaggle.com/c/rsna-intracranial-hemorrhage-detection}{Kaggle RSNA Intracranial Hemorrhage Detection challenge} \cite{rsna-intracranial-hemorrhage-detection}. 
We show that \gls{MAC} outperforms simple average methods, demonstrates competitive results to \xgboost for a fixed number of sub-models, and outperforms it when adding sub-models to the combination after the training phase.

The main contributions of our work are:
\begin{itemize}
\item We introduce a new ensemble method that is agnostic to the number and the details of the sub-models participating in the combination, and that can be optimized on specific tasks.
\item We describe an efficient and cost effective implementation of our method using a test case problem.
\item We show that our method outperforms existing averaging methods and is competitive with \xgboost. 
\item We demonstrate the robustness of our method to changes in the number and specifics of the  sub-models participating in the combination.
\item In order to allow full reproducibility of our results, we share our data and code at \href{https://doi.org/10.5281/zenodo.3648244}{https://doi.org/10.5281/zenodo.3648244}.
\end{itemize}

\section{Methods}
\label{sec:method}
As stated above, ensembles improve single models performance by averaging over many sub-model predictions. 
If each sub-model performs better than random, an average would enhance the correct prediction and suppress outliers (see \cite{when_networks_disagree, ensemble_methods_review_book} for details).

It is obvious that for a given ensemble of sub-models and a given evaluation metric different combination methods provide different results. 
Thus choosing the correct combination method is crucial for achieving the best performance. 
Alternatively, averaging based methods are convenient as they allow continuously adding models to the combination without the need to retrain the combination. 

Therefore, it is advantageous to incorporate the strength of averaging based methods with the necessity of optimizing for the best performing combination method.
We propose a framework that allows to search over a space of combination functions and optimize the combination with respect to the given evaluation metric. 
We design our framework such that the space of functions we search over is agnostic to the number of models participating in the combination.

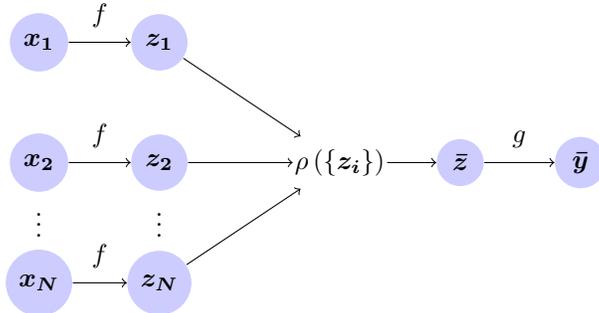
\begin{figure}[t]
	\centering
	\begin{tikzpicture}
	[scale=.8,auto=left,every node/.style={circle,fill=blue!20}]
	\node (n1) at (1,10) {$\boldsymbol{x_1}$};
	\node (n2) at (1,8) {$\boldsymbol{x_2}$};
	\node (n3) at (1,6) {$\boldsymbol{x_N}$};
	\node (n4) at (3,10) {$\boldsymbol{z_1}$};
	\node (n5) at (3,8) {$\boldsymbol{z_2}$};
	\node (n6) at (3,6) {$\boldsymbol{z_N}$};
	\node[circle,fill=none,inner sep=0pt] (n7) at (6,8) {$\rho\left(\left\lbrace\boldsymbol{z_i} \right\rbrace\right)$};
	\node (n8) at (8,8) {$\boldsymbol{\bar{z}}$};	
	\node (n9) at (10, 8) {$\boldsymbol{\bar{y}}$};
	\foreach \from/\to in {n1/n4,n2/n5,n3/n6}
	\draw[->] (\from) -- node[fill=none] {$f$} ++ (\to);
	\foreach \from/\to in {n4/n7,n5/n7,n6/n7}
	\draw[->] (\from) -- (\to);
	\draw[->] (n7) -- (n8);
	\draw[->] (n8) -- node[fill=none] {$g$} ++(n9);
	\node[fill=none] at (1,7.1) {\vdots};
	\node[fill=none] at (3,7.1) {\vdots};
	\end{tikzpicture}
	\caption{
		Diagram of \gls{MAC}. 
		The output predictions $\{\boldsymbol{x_1}, \boldsymbol{x_2}, \ldots, \boldsymbol{x_N}\}$ from $N$ sub-models are transformed by $f$ to the latent space. 
		The resulting $\{\boldsymbol{z_1}, \boldsymbol{z_2}, \ldots, \boldsymbol{z_N}\}$ latent space representations are combined using $\rho$.
		The latent space combination $\boldsymbol{\bar{z}}$ is then transformed from the latent space back to the output space using the function $g$ which provides the final prediction $\boldsymbol{\bar{y}}$.}
	\label{fig:generalcombination}
\end{figure}

The suggested \gls*{MAC} method is constructed from three functions. 
The first function, $f$, maps each sub-model predictions to a latent space in which the combination takes place. 
The second function, $\rho$, is a combination function, performed in the latent space and is chosen to be agnostic to the number and order of its input.
Finally, a third function, $g$, maps the combination result from the latent space back to the sub-model output space. 
An explicit form of the \gls*{MAC} method is given by

\begin{equation}
\boldsymbol{\bar{y}}\left(\left\lbrace \boldsymbol{x}_i \right\rbrace\right)=g\left(\rho\left( \left\lbrace f\left( \boldsymbol{x}_i \right) \right\rbrace \right)\right)
\label{eq:generalcombination}
\end{equation}

where $\boldsymbol{x}_i$ is the $i$-th sub-model predictions and $\boldsymbol{\bar{y}}$ is the final combination result. 
In our notation we do not restrict $\boldsymbol{x}_i$ to be a scalar.
It can be any feature vector a sub-model provides, as long as the feature space of all the sub-models is consistent. 
An illustration of the \gls*{MAC} structure is given in Figure~\ref{fig:generalcombination}.

The combination function $\rho$ is a predefined function that combines the latent space representations 
$\{\boldsymbol{z}_1, \boldsymbol{z}_2, \ldots, \boldsymbol{z}_N\}$, 
to a single representation that does not scale with the number of sub-models $N$. 
Some examples of possible combination functions are: averaging functions, voting functions, min, max and median. 

We can express different complex ensembling functions even by choosing $\rho$ to be a simple linear average. In this case, equation.~\eqref{eq:generalcombination} becomes:

\begin{equation}
\boldsymbol{\bar{y}}\left(\left\lbrace \boldsymbol{x_i} \right\rbrace\right)=g\left(\frac{1}{N}\sum_{i=1}^{N}f\left( \boldsymbol{x_i} \right)\right)
\label{eq:avgcombination}
\end{equation}

A simple average combination is a special case of \gls{MAC}, for which both $f$ and $g$ are just unity functions. 
However, choosing $\rho$ to be linear average does not limit the \gls{MAC} as one can get, for example, the harmonic and geometric averages by choosing the appropriate $f$ and $g$ functions - Harmonic averaging is modeled with $f$ and $g$ being the inverse function $\left(f(x)=g(x)=1/x\right)$, and the geometric averaging is achieved using $f(x)=\ln(x)$ and $g(x)=\exp(x)$. 
Thus the choice of $\rho$ to be a linear average is not a limiting factor and for the sake of simplicity we will use it for the rest of the paper. 
Since $\rho$ is agnostic to the number of sub-models, and $f$ operates similarly on every sub-model prediction, the total scheme of \gls{MAC} is agnostic the number of sub-models. 

The mapping functions $f$ and $g$ are parameterized and can be optimized to the specific task in hand. 
For example, both $f$ and $g$ can be manifested using predefined functions (polynomials, exponentials etc.). 
The functions parameters would then be determined using regression or some other estimation method. 
In general cases, a broader function space for $f$ and $g$ can be implemented using \gls*{DNN}s and optimized using \gls*{SGD}.
In such cases, it is reasonable to assume that a rich \gls*{DNN} structure (i.e. at least one hidden layer and a sufficient number of units) together with a simple combination function $\rho$ can replace many general combination functions. 
The latent space dimension can also be a hyper-parameter of the problem and can be optimized using some search method (e.g. grid-search). 
Moreover, the latent space dimensions can be different than the dimension of each sub-models prediction.

\section{Experiments and Results}
\label{sec:experiment}
\subsection{Experiments}
\label{sec:experiment_description}

\begin{figure}[t]
	\includegraphics[width=\columnwidth]{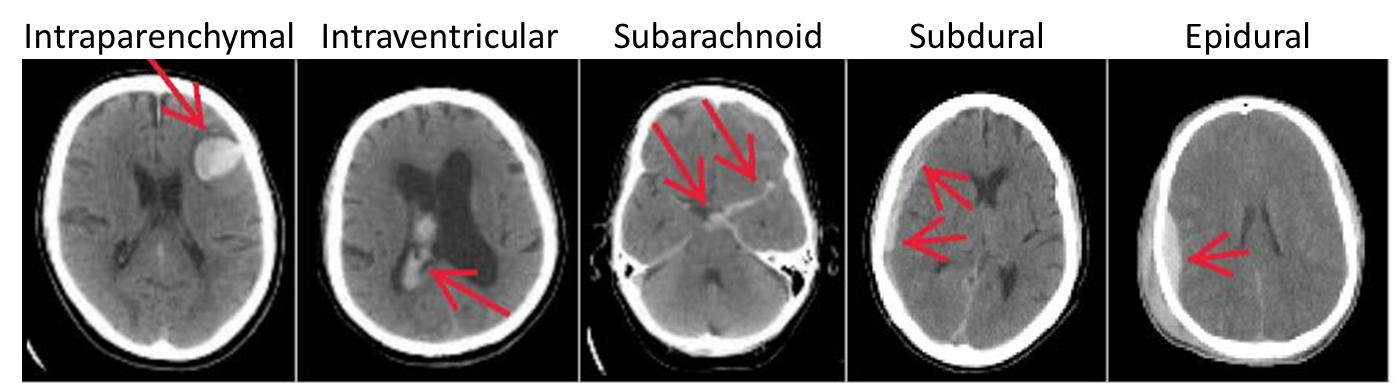}
	\vskip -0.05in
	\caption{
		Example slices of brain \gls{CT} scans classified to sub-types of hemorrhage. Image via \href{https://www.kaggle.com/c/rsna-intracranial-hemorrhage-detection/overview/hemorrhage-types}{https://www.kaggle.com/c/rsna-intracranial-hemorrhage-detection/overview/hemorrhage-types}
		}
	\label{fig:rsna_challenge}
	\end{figure}

We analyze the performance of \gls{MAC} on a test case - the \href{https://www.kaggle.com/c/rsna-intracranial-hemorrhage-detection}{RSNA Intracranial Hemorrhage Detection challenge} hosted by Kaggle \cite{rsna-intracranial-hemorrhage-detection}. 
The objective of the challenge was to detect acute intracranial hemorrhage and its sub-types. 
The training dataset contained over $750$K $2D$ slices of brain \gls{CT} scans. 
Each slice was independently labeled with either none or any of the five following classes: Epidural, Intraparenchymal, Intraventricular, Subarachnoid and Subdural hemorrhage. 
Slices labeled with at least one of the hemorrhage sub-types were also labeled with class "any" which indicates that a hemorrhage exists in the image. 
Example slices and labels are demonstrated in Figure~\ref{fig:rsna_challenge}.

Participating solutions were evaluated using a binary cross-entropy loss averaged over the 6 classes with the "any" label weighting twice the other sub-types. 
The challenge was a two-stage competition\footnote{\href{https://www.kaggle.com/two-stage-frequently-asked-questions}{https://www.kaggle.com/two-stage-frequently-asked-questions}} 
and on each stage, a different test set of over $120$K new slices was released.
Once the second stage starts, the ground-truth labels of the first stage test set are released and the first stage test set becomes a part of the second stage training set.

To address this challenge, we ensembled a total of $460$ sub-models to predict the class probability of hemorrhage sub-types in every image. 
These models were hand tailored versions of pre-trained 2D classification networks such as \textit{Se\_Resnext50\_32x4d}, all trained using the weighted binary cross-entropy loss defined earlier. 
The sub-models varied in their input structure (e.g. number of input slices), hyper parameters (normalization, augmentations, optimizer etc.) and subsets of the training set (5 fold cross validation training scheme). 
$310$ sub-models were trained on the training data of the first stage, and the remaining $150$ were trained on the test data of the first stage together with the training set. 
The individual models scored between $0.051 - 0.060$ on the withheld second stage test-set. 
A simple average over the $310$ and $460$ sub-models predictions scored $0.04974$ and $0.04984$, respectively. 
Once establishing the baseline results on this task, we turn to evaluate the performance of \gls*{MAC} and compare it to \xgboost  \cite{xgboost}.

Both \gls*{MAC} and \xgboost require supervised training.
We generated predictions of each sub-model on the full data of over $720K$ images.  
Due to computational constraints (acquiring the full predictions over the training-set data), we generated predictions of only $N=310$ sub-models and used these predictions for training. 
The performance was evaluated on the second stage test-set using predictions from the same $310$ sub-models. 
Since the \gls{MAC} is agnostic to the number of sub-models, we could analyze the score improvement as we added predictions from additional sub-models to the combination.

Our code is implemented in PyTorch and can be found, together with
the predictions of all sub-models on the training and test data, at 
\url{https://doi.org/10.5281/zenodo.3648244}. 
We used NVIDIA Titan RTX GPU for training and inference of the MAC, training until convergence took about 2 days.

\subsection{Ensemble via \gls{MAC}}
\begin{figure}[t]
	\begin{center}
	\includegraphics[width=\columnwidth]{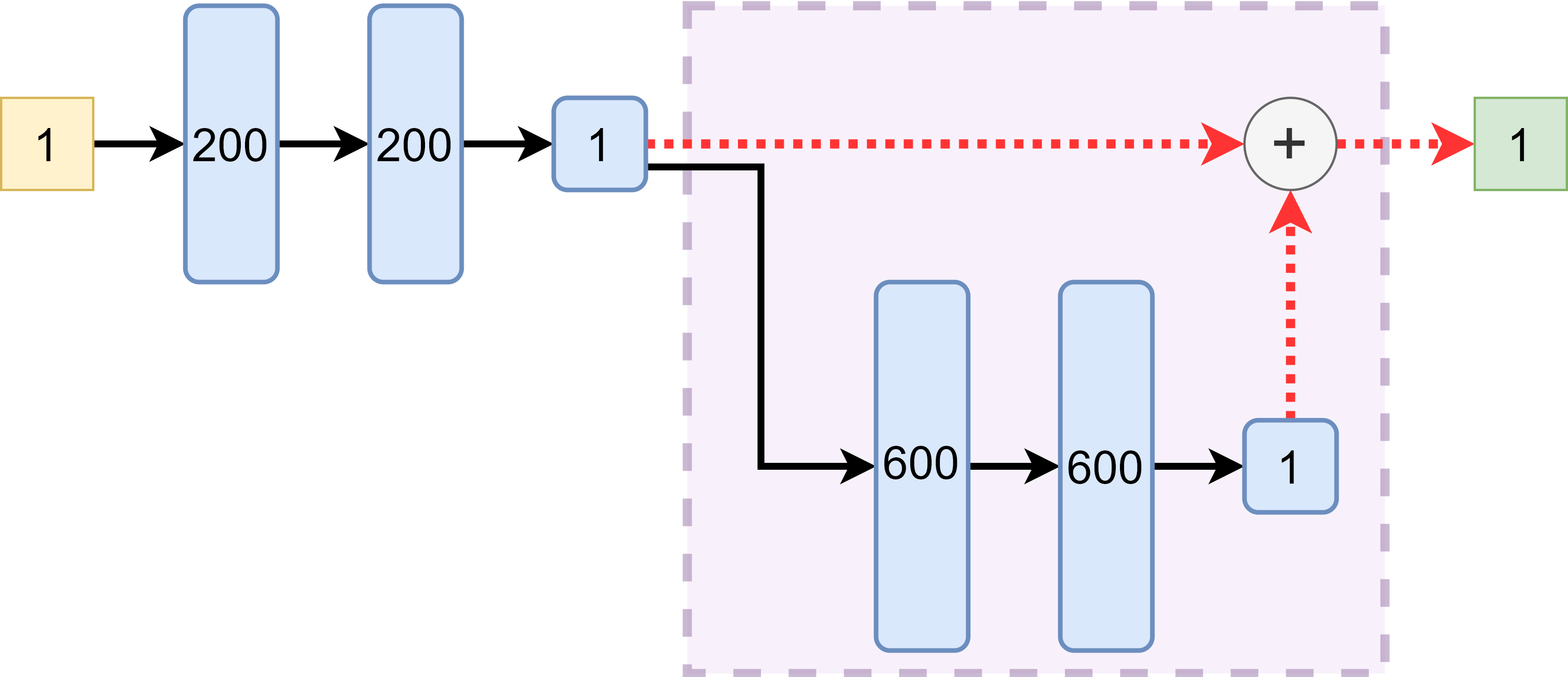}
	\end{center}
	\caption{
		\gls{DNN} architecture of $f$ and $g$ for the task of acute intracranial hemorrhage detection.
        Each (blue) block represents a fully connected layer of \gls{RELU}s, with the number of units in each layer reported in it.
        Solid black arrows represent connections to the next layer, red dotted arrows represent identity.
        The purple square contains the residual block. 
		The output layer (green) has no activation in $f$ and a Sigmoid activation in $g$.
		}
	\label{fig:network}
	\end{figure}
	
We implement the functions $f$ and $g$ as \gls*{DNN}s and, for simplicity share the same architecture. 
In order to minimize the learned parameter space and improve convergence, we use the same $f$ and $g$ for all $6$ classes. 
The dimensions of the input and output spaces are the same and since we combine each class individually they are both equal $1$, i.e. the class probability (see Figure \ref{fig:network}). 
For simplicity, we fix the latent space to one dimension, and the combination function $\rho$ to a linear average. 
Equation \eqref{eq:model} summaries our choice of functions:

\begin{equation}
\begin{aligned}
&f: [0, 1]\rightarrow\mathds{R} \\
&\rho\left(\left\lbrace z_i \right\rbrace\right)=\frac{1}{N}\sum_{i=1}^{N} z_i \\
&g: \mathds{R}\rightarrow[0, 1] \\
\end{aligned}
\label{eq:model}
\end{equation}

The \gls*{DNN}s have the following architecture, illustrated in Figure \ref{fig:network}: 
The input is processed by two fully connected layers with $200$ \gls*{RELU}s, which is sufficient to represent a complex analytical
transformation function. 
Then, forwarded to one residual block in order to ease convergence and increase accuracy \cite{he2015deep_resnet}. 
The residual block contains two fully connected layers with $600$ \gls*{RELU}s. 
The output of the residual block is summed together with its input to produce the output of the network. 
The last layer of $f$ has no activation since $f$ transforms input to the latent space $\mathds{R}$. 
On the other hand, $g$ has a Sigmoid activation in its last layer to represent the class probability.
This architecture allows an end to end training of the full \gls*{MAC}.

Training the MAC is a supervised learning problem with the same objective function as each of the sub-models.
We split the first-stage training dataset, to $80\%, 10\%, 10\%$ stratified splits for train, validate and test sets, respectively. 
In each iteration we randomly select $0.8N$ of the sub-models for the combination. 
This is done to encourage sub-model invariance.
The training loss is the binary cross-entropy function and the ADAM \cite{adam} optimizer variant of \gls*{SGD} is used with a constant learning rate of $10^{-3}$. 
We train the \gls*{MAC} for $45-60$ epochs with a batch size of $500$ samples until validation loss stabilizes.

\subsection{Ensemble via \xgboost }
\begin{table}[t]
	\caption{Hyper parameters of XGBoost model}
	\label{tab:xgboost-table}
	\begin{center}
	\begin{small}
	\begin{sc}
	\begin{tabular}{lr}
	\toprule
	Parameter & Value \\
	\midrule
	$\alpha$ & $0.5$ \\
	$\gamma$ & $0.5$\\
	max depth & $3$ \\
	min child weight &$6$ \\
	n estimators &$1000$\\
	$\eta$ & $0.05$ \\
	objective & binary:logistic \\
	eval metric & logloss\\
	subsample & $0.5$ \\
	colsample bytree & $0.8$ \\
	\bottomrule
	\end{tabular}
	\end{sc}
	\end{small}
	\end{center}
	\vskip -0.1in
\end{table}

We compare the \gls{MAC} performance to \xgboost \cite{xgboost}, one of the leading ensembling methods today.
We train 6 models - one per class. Every model receives the predictions from all classes as input, resulting in $6 \times 310 = 1860$  input features. 
Thus, the \xgboost could potentially exploit the correlations between class predictions to better predict the correct class.
The training dataset is split to $80 \%  - 20\%$ train and validation sets. 
Hyperparameters are tuned through grid search and are listed in table \ref{tab:xgboost-table}.
We train the models for $1000$ steps with an early stopping rule after 2 rounds of no improvement in the validation loss. 

\subsection{Results}
\begin{table}[t]
	\caption{Comparison of combination methods}
	\label{tab:results_comparison}
	\begin{center}
	\begin{small}
	\begin{sc}
	\begin{tabular}{lcc}
	\toprule
	Combination & $310$ models score& $460$ models score\\
	\midrule
	Average & $0.04974$ & $0.04984$ \\
	\xgboost  & $0.04657$ & N/A \\
	\gls*{MAC} & $0.04662$ & $0.04644$\\
	\bottomrule
	\end{tabular}
	\end{sc}
	\end{small}
	\end{center}
\end{table}

A summary of the results is given in table \ref{tab:results_comparison}. 
We start by comparing the results obtained from the combination of $N=310$ sub-models. 
In this case, \xgboost scored $0.04657$ on the weighted binary cross entropy score described earlier (see section \ref{sec:experiment_description}), and similarly, \gls*{MAC} scored $0.04662$, both improving dramatically the baseline results of a simple linear average 
(corresponding to an increase of 17 ranks on the challenge leaderboard). 
We were able to improve the \gls*{MAC} score to $0.04644$ by using predictions from all the $460$ sub-models we had without retraining. 
This procedure is impossible with a boosting method like \xgboost  without retraining on all sub-models. 

\subsection{Analyzing \gls*{MAC}'s performance}

\begin{figure}[t]
\begin{center}
\includegraphics[width=\columnwidth]{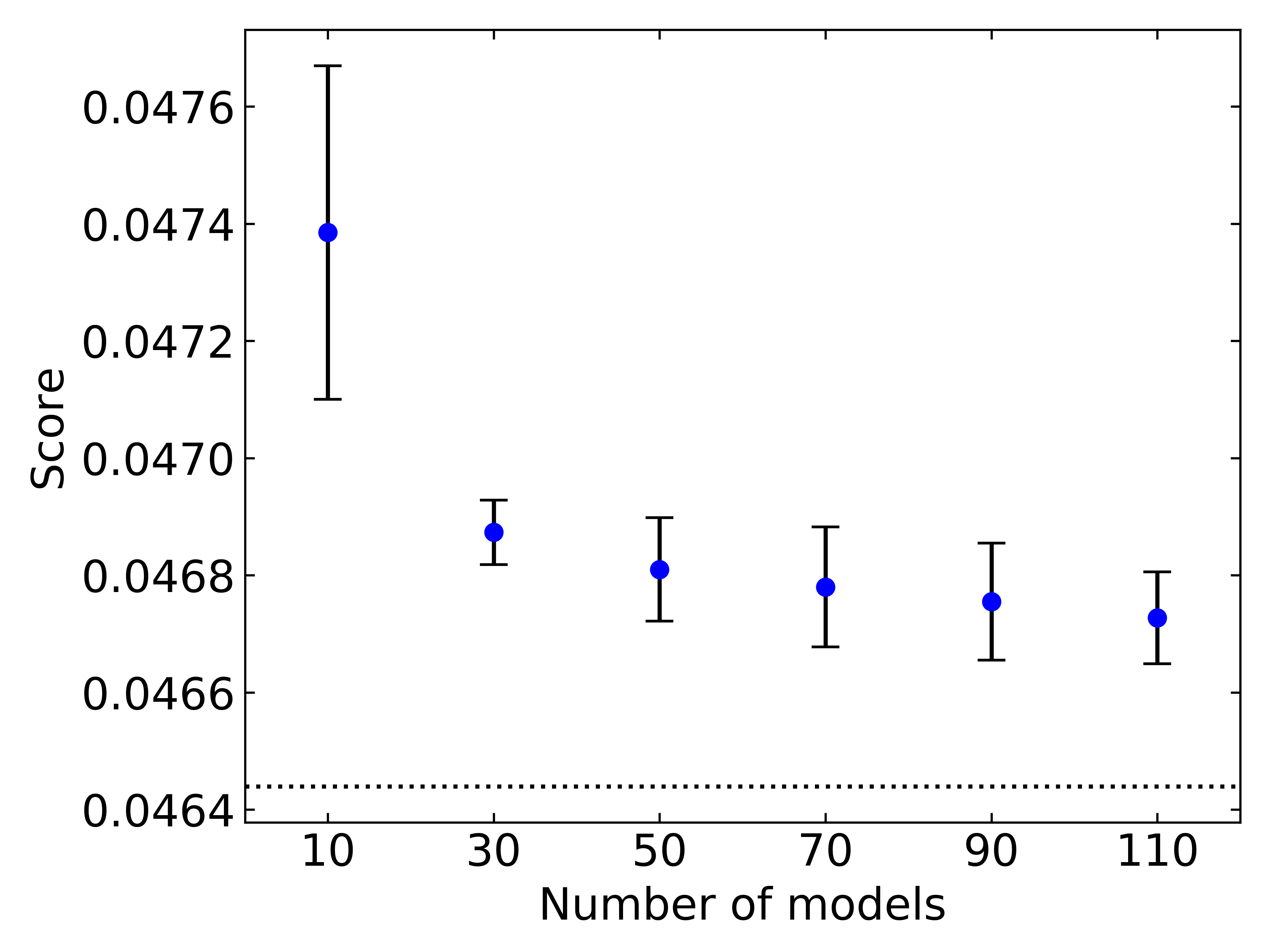}
\vskip -0.1in
\caption{
	Score vs. number of sub-models participating in the combination via \gls*{MAC} when trained on $310$ sub-models (lower is better). 
	Each circle represents the average score of $4$ experiments on different groups of sub-models and the error bars indicate the standard deviation. 
	The black dotted line indicates the best score achieved by predicting with $460$ sub-models
	}
\label{fig:mac_improvement}
\end{center}
\end{figure}

\begin{figure}[t]
	\begin{center}
	\includegraphics[width=\columnwidth]{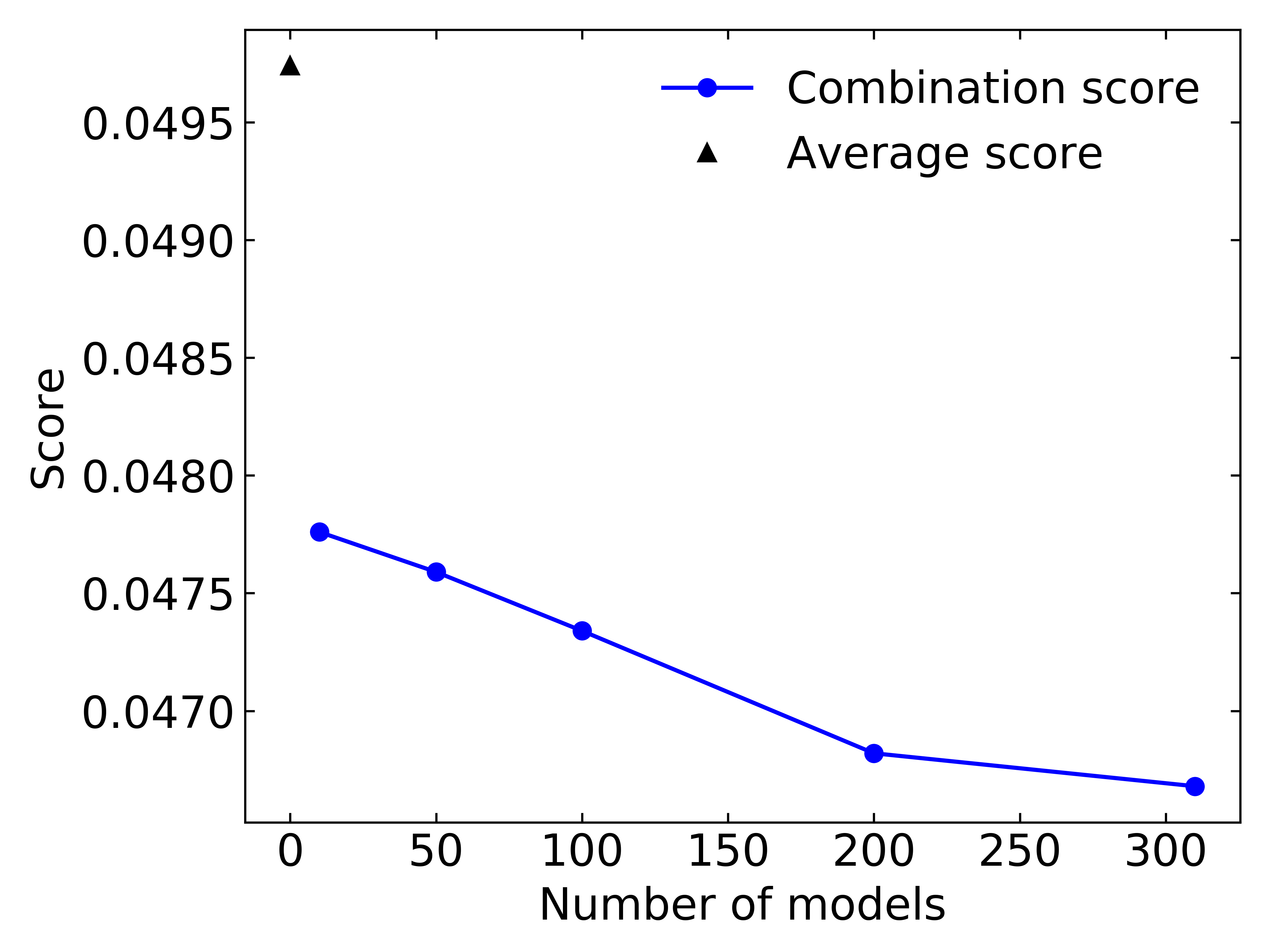}
	\caption{
		Score obtained by prediction with $460$ sub-models vs. number of sub-models (10, 50, 100, ..) used in \gls*{MAC} training (lower is better).
		Here, experiments cannot be reproduced with different \textbf{uncorrelated} samples. 
		Therefore, each blue point represents a single experiment. 
		The black triangle indicates the score of a simple average of $460$ models
		}
	\label{fig:combination_score_vs_trained_models}
	\end{center}
\end{figure}

While it may be straight-forward that adding sub-models improves the ensemble results, we investigate the performance of a 
trained \gls*{MAC} with a varying number of sub-models participating in the prediction. 
Therefore, we combine predictions of $10, 30, 50, 70, 90$ and $110$ sub-models using the \gls*{MAC} trained on $310$ sub-models. 
We repeat this experiment $4$ times with different sets of sub-models and measure the score each time. 
The results are presented in Figure. \ref{fig:mac_improvement}.
Each point corresponds to the mean of the $4$ scores obtained for a fixed number of sub-models and the error bars indicate 
the standard deviation. 
The black dotted line marks the best score obtained by predicting with all $460$ models. 
Note that predicting with only $10$ models already improves the simple average score, and as expected, 
adding sub-models to the prediction combination improves the score. 

As mentioned above, training any combination model requires obtaining predictions of the sub-models over the full train set. When the train set
is extremely large (which is often the case), doing so for many sub-models may be exhaustive. One way to overcome this challenge when using
\gls{MAC}, due to its invariance to sub-models, is to obtain predictions on the train set from a small subset of sub-models, and use that for training.
During prediction the combination can still use all sub-models to further improve results. To demonstrate this ability, 
we test the influence of the number of models used to train \gls*{MAC} on the score obtained when predicting on all $460$ models. 
We trained \gls*{MAC} on $10, 50, 100, 200$ and $300$ sub-models and evaluated their performance.
The results, demonstrated in Figure. \ref{fig:combination_score_vs_trained_models}, indicate that training \gls*{MAC} on only $10$ 
models already improves the average score. 
Again, adding sub-models to the training indeed improves performance.

\subsection{Analyzing \gls*{MAC}'s functions}

\begin{figure}[!ht]
\begin{center}
\includegraphics[width=\columnwidth]{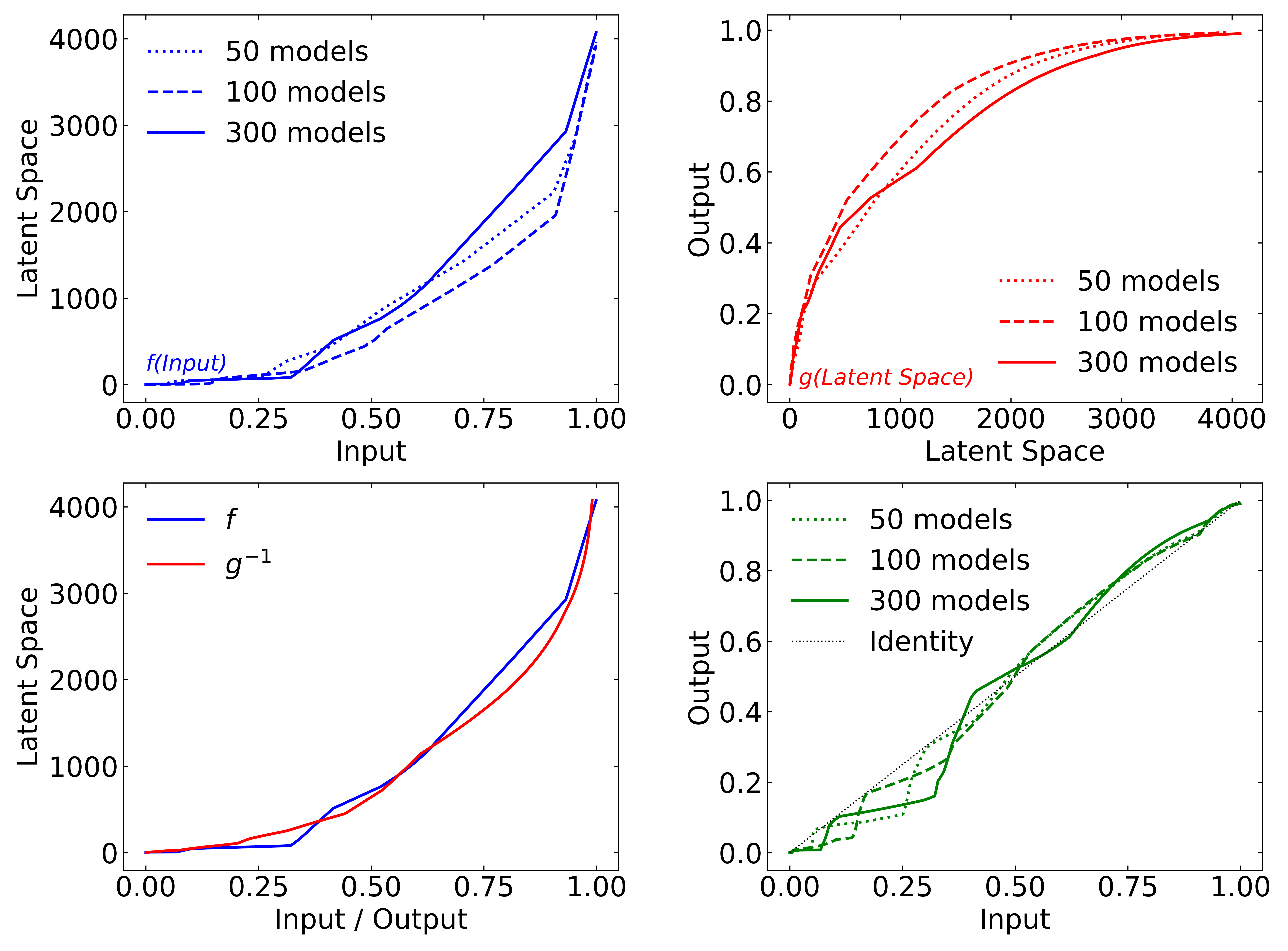}
\caption{
	Evaluations of the trained \gls{DNN}s $f$ and $g$ in various cases: 
	Top left: $f$ (in latent space arbitrary units) as a function of \gls*{MAC}s input, trained on 50 (dotted line), 100 (dashed line) and 300 (solid line) sub-models, 
	all demonstrating similar behavior. 
	In order to plot all lines on the same axes, each curve was normalized by a constant value. 
	Top right: same for $g$ as a function of $f$ in latent space. 
	Bottom left: $f$ and and $g^{-1}$ trained on $300$ models, plotted on the same axes to demonstrate their similarity. 
	Bottom right: \gls{MAC}s output as a function of \gls{MAC}s input when trained on 50 (dotted line), 100 (dashed line) and 300 (solid line) sub-models. 
	The identity curve is plotted in a dotted black line.}  
\label{fig:mac_f_g_functions}
\end{center}
\end{figure}

To gain insight into how \gls*{MAC} works we evaluate the behavior of the learned functions $f$ and $g$.
First, we examine $f$ by plotting its value for the input range of $[0, 1]$ (recall that the input to $f$ is a
the probability belonging to some class as predicted by a sub-model). 
The top-left panel in Figure. \ref{fig:mac_f_g_functions} demonstrates $f$'s output in latent space arbitrary units as a 
function of its input, for \gls*{MAC}s trained on $50$, $100$ and $300$ models. 
The general behavior is similar regardless of the number of sub-models used in training, 
which implies that the general optimal solution for this specific challenge is stable and highly non-linear - 
strongly enhancing the input as it approaches $1$. 
i.e. positive predictions are strongly enhanced and therefore weighted higher than negative ones. 
Note, that while regular weighted average methods assign fixed weights to every sub-model for all predictions 
coming from that sub-model, here, the weighing is performed per prediction value regardless of the sub-model it came from. 

We now turn to examine $g$. To this end, we wish to separate the effect of $\rho$ on the \gls{MAC} performance. Therefore, we treat each 
input in the range of $0-1$ as a single input to \gls{MAC}. Then, $\rho$ has no effect on its input and 
Equation.\eqref{fig:generalcombination} reduces to $g(f(x))$. 
Results for $g$ are given in the top right panel of Figure. \ref{fig:mac_f_g_functions}. Similar to $f$, $g$ also shows stability with
respect to the number of sub-models used in training. 

In the bottom panels of Figure. \ref{fig:mac_f_g_functions} we analyze the similarity between $f$ and $g^{-1}$ - the 
inverse function of $g$, meaning, the output of $g$ as a function of it's input - $f(x)$. 
On the left, $f$ and $g^{-1}$ are plotted on the same axes, demonstrating their similarity which implies that $f \approx g^{-1}$,  
this was not predefined by the architecture but rather learned through training. 
On the bottom right panel, the output of $g$ is plotted as a function of the input to $f$ and is compared to the identity curve (black dotted line). 
When examining the curves for different numbers of sub-models participating in the combination, it is easy to see that the general behavior is close to the identity function. 
A closer look reveals that all the curves behave similarly in the sense that they are consistently above identity for inputs above $0.5$ while below identity for inputs below $0.4$. 
By slightly enhancing averaged positive predictions and suppressing negative ones, $g$ boosts the final averaged prediction. This property demonstrates again the nontrivial optimal solution found by \gls*{MAC}.

\section{Discussion}
\label{sec:discussion}

\begin{figure}[t]
\centering
\begin{tikzpicture}
[scale=.8,auto=left,every node/.style={circle,fill=blue!20}]
\node[align=center] (n1) at (1,10) {$\boldsymbol{x_1}$\\$w_1$};
\node[align=center] (n2) at (1,8) {$\boldsymbol{x_2}$\\$w_2$};
\node[align=center] (n3) at (1,6) {$\boldsymbol{x_N}$\\$w_N$};

\node[align=center] (n4) at (3,10) {$\boldsymbol{z_1}$\\$w_1$};
\node[align=center] (n5) at (3,8) {$\boldsymbol{z_2}$\\$w_2$};
\node[align=center] (n6) at (3,6) {$\boldsymbol{z_N}$\\$w_N$};

\node[circle,fill=none,inner sep=0pt] (n7) at (6,8) {$\rho=\frac{\sum_{i=1}^{N} w_i \boldsymbol{z_i}}{\sum_{i=1}^{N} w_i}$};
\node (n8) at (8,8) {$\boldsymbol{\bar{z}}$};
\node (n9) at (10, 8) {$\boldsymbol{\bar{y}}$};

\foreach \from/\to in {n1/n4,n2/n5,n3/n6}
{
	\draw[->,transform canvas={yshift=-2}] (\from) -- node[fill=none] {$f$} ++ (\to);
	\draw[->,transform canvas={yshift=2}] (\from) -- node[fill=none] {} ++ (\to);
}
\foreach \from/\to in {n4/n7,n5/n7,n6/n7}
{
	\draw[->,transform canvas={yshift=-2}] (\from) -- (\to);
	\draw[->,transform canvas={yshift=2}] (\from) -- (\to);
}
\draw[->] (n7) -- (n8);
\draw[->] (n8) -- node[fill=none] {$g$} ++(n9);

\node[fill=none] at (1,7.1) {\vdots};
\node[fill=none] at (3,7.1) {\vdots};

\end{tikzpicture}
\caption{
	Combination of $N$ models using weighted sum in latent space, where the weights are given in the feature vector of each sub-model.
	}
\label{fig:weighted_mac}
\end{figure}
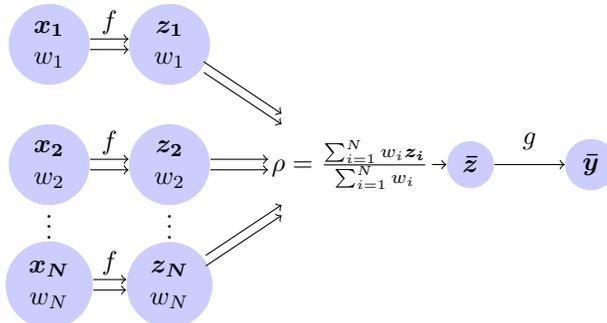

We introduce a new framework, \gls*{MAC}, for building an ensemble of sub-models, for which, the task of finding optimal ensembles 
is treated as a supervised learning problem.
The major advantage of the \gls*{MAC} framework is the flexibility in changing the sub-models (learners) in the combination during inference without retraining the combination function. 
We present a simple implementation of this approach using \gls*{DNN}s and compare its performance to a simple average method and to \xgboost. 
We also demonstrate that \gls*{MAC} can outperform \xgboost  by adding more sub-models during prediction. 

By further analyzing \gls*{MAC}s performance when changing the number of sub-models participating during training and inference, we demonstrate that while the score improves as more sub-models are added, an improvement with respect to simple average can be obtained even when using a small number of models in the combination. 
This advantage can be used to economize computational resources during training and inference. 
Furthermore, we show that the functions $f$ and $g$ are highly non-linear and therefore, enable \gls*{MAC} to produce complex combinations which are infeasible with average methods.
It is worth noting that another advantage of the implementation of \gls*{MAC} using \gls*{DNN}s when the sub-models are also \gls*{DNN}s is that the training of the \gls*{MAC} can be composed together with the training of the sub-models in an end-to-end manner. 

Since $f$ and $g$ transform from one space to the other, they may be the inverse of each other. 
One way to encourage this behavior is by minimizing the distance between the output of $g$ and the input of $f$, 
similar to auto-encoders \cite{autoencoder}. 
However, we suggest instead to optimize \gls*{MAC} with the same loss designed for the task in hand. 
This will optimize $f$ for which $\rho$ performs best and at the same time optimize $g$ such that the 
combined prediction is boosted towards the correct class.

There are many ensembling and averaging methods (e.g. softmax, p-norm, averaging, etc.) which can be used to generate predictions from multiple sub-models while being agnostic to the number of inputs. 
However, one must choose between them or exhaustively search for the best method. 
The \gls{MAC}, on the other hand, provides (when using \gls{DNN}s) a framework for finding an agnostic ensembling method without 
limiting the search to a small number of predefined functions.

In our work, we focus on combining sub-model predictions. However, in some cases it may be beneficial to combine sub-model features instead. 
The \gls{MAC} framework is able to combine any type of the sub-model outputs: features, logits and predictions. 
We leave the investigation of the advantages of each of these cases to future work. 
Moreover, more complex models can fit into the \gls*{MAC} approach as well. 
For example, the combination function can be generalized to weight the sub-models based on some features (similar to weighted average), keeping its result agnostic to the number of sub-models. 
A way to achieve a weighted sum is to separate the feature vector of each sub-model to features we want to combine and weight-features. 
A simple example is illustrated in Figure \ref{fig:weighted_mac}.

There are cases for which there are several groups of sub-models instead of a single one and we suspect that they should not be combined using the same metric.
An example for such a case would be several n-folds, where each n-fold has a different training set or trained using a different loss function. 
In that case a hierarchical combination may be used where each fold is combined first. 
The resulting combination is then treated as a new sub-model, and can be further combined by repeating the combination procedure. 

\pagebreak
\bibliographystyle{ieeetr}
\bibliography{references}

\end{document}